\documentclass[letterpaper, 10 pt, conference]{ieeeconf}

\IEEEoverridecommandlockouts                              % This command is only needed if 
\usepackage[utf8]{inputenc}
\usepackage{amsmath}
\usepackage{mathrsfs}
\usepackage{tcolorbox}
\usepackage{graphicx}
\usepackage{subfig}
\title{\LARGE \bf A Quasi-centralized Collision-free Path Planning Approach for Multi Robot Systems}

\author{Rohith G., \textit{Member, IEEE}
  and Madhu Vadali$^{*}$, \textit{Member, IEEE}% <-this % stops a space
\thanks{*Corresponding author}
\thanks{Rohith G. {\tt\footnotesize (rohith044@gmail.com)}, SMART Lab, Mechanical Engineering Discipline, Indian Institute of Technology Gandhinagar.}
\thanks{Madhu Vadali {\tt\footnotesize (madhu.vadali@iitgn.ac.in)}, SMART Lab, Jointly with Mechanical Engineering and Electrical Engineering Disciplines, Indian Institute of Technology Gandhinagar.}
\thanks{\textit{This work has been submitted to the IEEE for possible publication. Copyright may be transferred without notice, after which this version may no longer be accessible}.}

}

\begin{document}

\maketitle
\thispagestyle{empty}
\pagestyle{empty}


\begin{abstract}
This paper presents a novel \textit{quasi-centralized} approach for collision-free path planning of multi-robot systems (MRS) in obstacle ridden environments. A new \textit{formation potential fields} (FPF) concept is proposed around a \textit{virtual agent}, located at the center of the formation which ensures self-organization and maintenance of the formation. The path of the virtual agent is centrally planned and the robots at the minima of the FPF are forced to move along with the virtual agent. In the neighborhood of obstacles, individual robots selfishly avoid collisions, thus marginally deviating from the formation. The proposed quasi-centralized approach introduces formation flexibility into the MRS, which enables MRS to effectively navigate in an obstacle ridden work space. Methodical analysis of the proposed approach and guidelines for selecting the FPF are presented. Results using a candidate FPF are shown that ensure a pentagonal formation effectively squeezes through a narrow passage avoiding any collisions with the walls.

\end{abstract}



% \fi
\IEEEpeerreviewmaketitle

\section{Introduction}
Multi-robot systems (MRS) embrace the idea of multiple robots working and navigating cooperatively to accomplish a specific task. It is common for the robots to move/execute tasks in a rigid/flexible formation depending on the application. While navigating through obstacle ridden fields, the robots should plan collision-free trajectories while maintaining the formation with varying levels of strictness, thus adding flexibility to formation. This flexibility could involve scaling \cite{zhao2018affine}, deforming \cite{rampinelli2010embedding}, and splitting \cite{olfati2006flocking} the formation to avoid obstacles. Multiple methods, viz., behavioral-based control methods \cite{balch1998behavior}, leader-follower approaches \cite{honig2016formation}, virtual structure approaches \cite{roy2018multi}, and artificial potential field based approaches \cite{schneider2003potential, song2002potential} have been used for the path planning and formation control of multi-robot systems. 

Behavioral-based control approaches, such as flocking and schooling, suffer from convergence issues \cite{olfati2004consensus}. The leader-follower approach becomes  complex for a large number of robots because of the dependency of the formation shape on the number of leaders \cite{cheah2009region}. Virtual structure approaches assume the formation to be confined within a geometric envelope, which is then treated as a single entity \cite{cheah2009region,tan1996virtual}. While this approach makes the planning problem simpler, it limits the flexibility \cite{roy2018multi} to find collision-free paths. Finally, artificial penitential field (APF) approaches, originally proposed for a single robot path planning \cite{khatib1985real}, focus more on establishing and maintaining a formation without colliding with each other \cite{roy2018multi}. A commonality among these approaches is that the paths are \textit{centrally} planned. APF methods with flexible formation for navigating an obstacle-ridden environment have rarely been explored \cite{roy2018multi, schneider2003potential}.  
%For a task that requires a group of robots establishing and maintaining a formation while navigating through an obstacle field, the collision-free individual robot paths should be planned considering the formation constraints also. 

This paper attempts to address the aforementioned limitations associated with the existing methods and proposes a APF based  \textit{quasi-centralized} path planning approach. Conventional APF planning approaches as in  \cite{khatib1985real,garibotto1991path} are used for planning the collision-free paths. However, instead of planning the paths for individual robots as in  \cite{schneider2003potential,song2002potential}, first a path for a \textit{virtual agent}, located at the center of the formation, is centrally planned. Subsequently, the paths for each robot are planned in  a decentralized and modular manner. The concept of a virtual agent is adopted from Leonard et al. \cite{leonard2001virtual}. While Leonard et al. create multiple virtual agents and centrally plan and control the formation, the proposed approach utilize only one virtual agent at the center of the formation. A novel potential field concept, called \textit{formation potential fields} (FPF), is used for self-organization and closely maintain the formation during navigation around the virtual agent. 

 In a pure centralized approach, all robots move so as to minimize the entire formation potential, and in a pure decentralized approach, the robots decide their on paths, decreasing their own potential, giving less regard to the overall formation potential. In the proposed quasi-centralized approach, the robots follow centrally planned path until they encounter an obstacle. In the neighborhood of the obstacles, the robots move to minimize their potential and avoid collision in a \textit{selfish} (decentralized) manner, making the whole formation to adjust and deform. 

\section{Problem Definition}
Consider a multi-robot system (MRS) of $N$ identical robots, in an obstacle ridden environment, represented by their position $\mathcal{\underline{Q}} = \{\underline{q}_1, \underline{q}_2,...,\underline{q}_N \}$ in $R^{2N}$ space. Here, $\underline{q}_i=[x_i, y_i]^T$ is the position  $i^{th}$ robot. In this paper, a \textit{formation} is defined as a $N$ sided regular polygon with each robot occupying the $N$ vertices.  The problem is to plan paths for the robots from an initial position $\mathcal{\underline{Q}}_i$ to a final position $\mathcal{\underline{Q}}_f$ such that,
\begin{itemize}
\item they self-organize into a desired formation
\item they reach the $\mathcal{\underline{Q}}_f$ in a desired formation
\item they neither collide with each other nor with the obstacles
\item they are not required to strictly maintain the formation
\end{itemize}
Without loss of generality, it is assumed that the initial and final desired formations are the same. 

\section{Formation Potential Fields}
Self-organization of the robots into a desired formation and the motion of the formation thereof is achieved using a novel concept called formation potential fields (FPF), $U_v$. These potentials fields, generated by a \textit{virtual agent} located at the center of the formation, $\underline{q}_{v}$, are designed to serve two purposes. One, they attract the robots to settle at a desired distance $R$, from virtual center. And two, they drag the robots along, when the virtual agent changes its location. Thus, the paths of the $N$ robots is \textit{centrally} decided by these FPF in absence of obstacles.

To achieve these goals and to maintain a stable formation, a FPF must have the following properties:
\begin{enumerate}
\item The global maximum is at the center, $\underline{q}_{v}$, of the formation.
\item The global minima are at a distance $R$ from the center and symmetric about the center. 
\item The function must be monotonically decreasing from $\underline{q}_{v}$ to $R$ and monotonically increasing from $R$ to $\infty$.
\end{enumerate}

While the first property guarantees that the robots do not cross the center and risk collisions with each other, the second and the third properties together ensure self-organization at $R$ from arbitrary initial conditions. Finally, in order to avoid collisions with each other, each individual robot must be associated with a repulsive field. Therefore, the $i^{th}$ robot experiences an attractive pull by the virtual agent and repulsive pushes by the remaining $N-1$ robots around it. Thus, in an obstacle-free environment, the total potential acting on the $i^{th}$ robot is,\begin{equation}
    U_i = U_{v} + \sum_{i \neq j}^{N} U_{ij},
\end{equation}where, $U_{ij}$, is the repulsive potential field experienced by the $i^{th}$ robot due to the $j^{th}$ one. The force acting on the $i^{th}$ robot, $\underline{F}_i$ is simply calculated as the negative gradient of the field or $-\nabla_i U_i$.

\begin{figure}[!h]
\centering
\subfloat[Sample FPF.]{\includegraphics[width=0.24\textwidth]{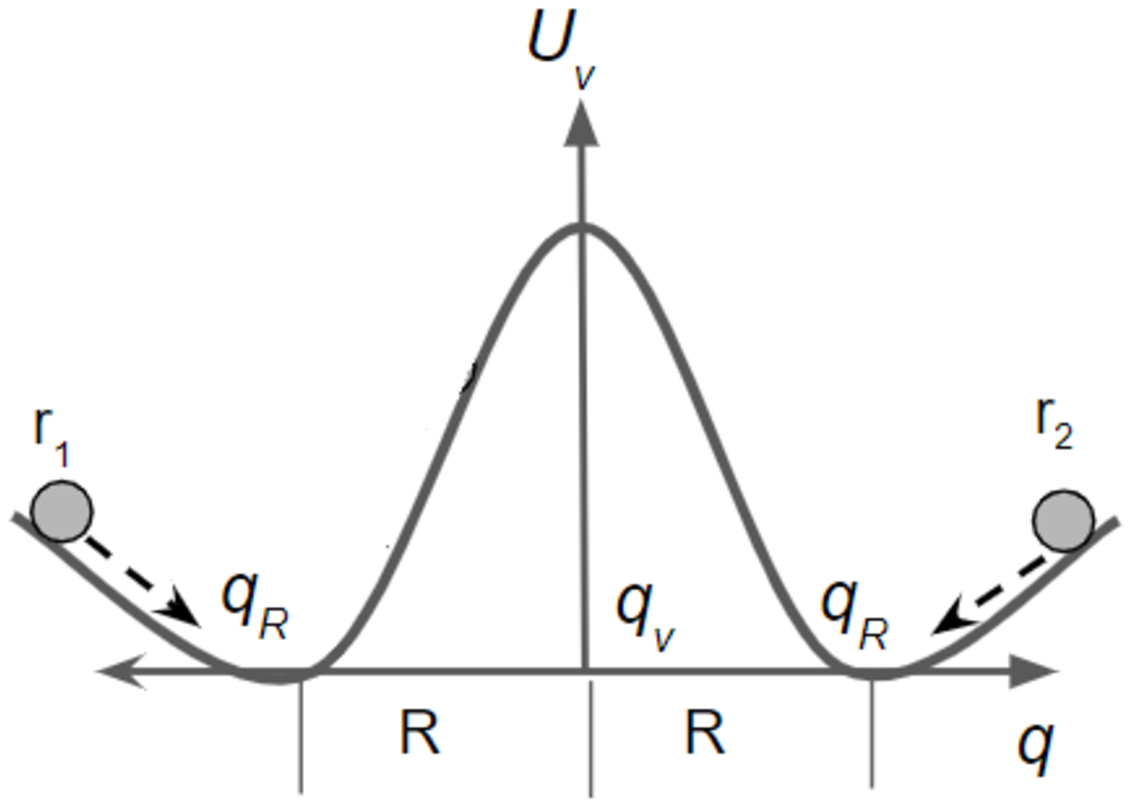}%
\label{F0}}
\hfil
\subfloat[A typical formation based on the proposed approach.]{\includegraphics[width=0.24\textwidth]{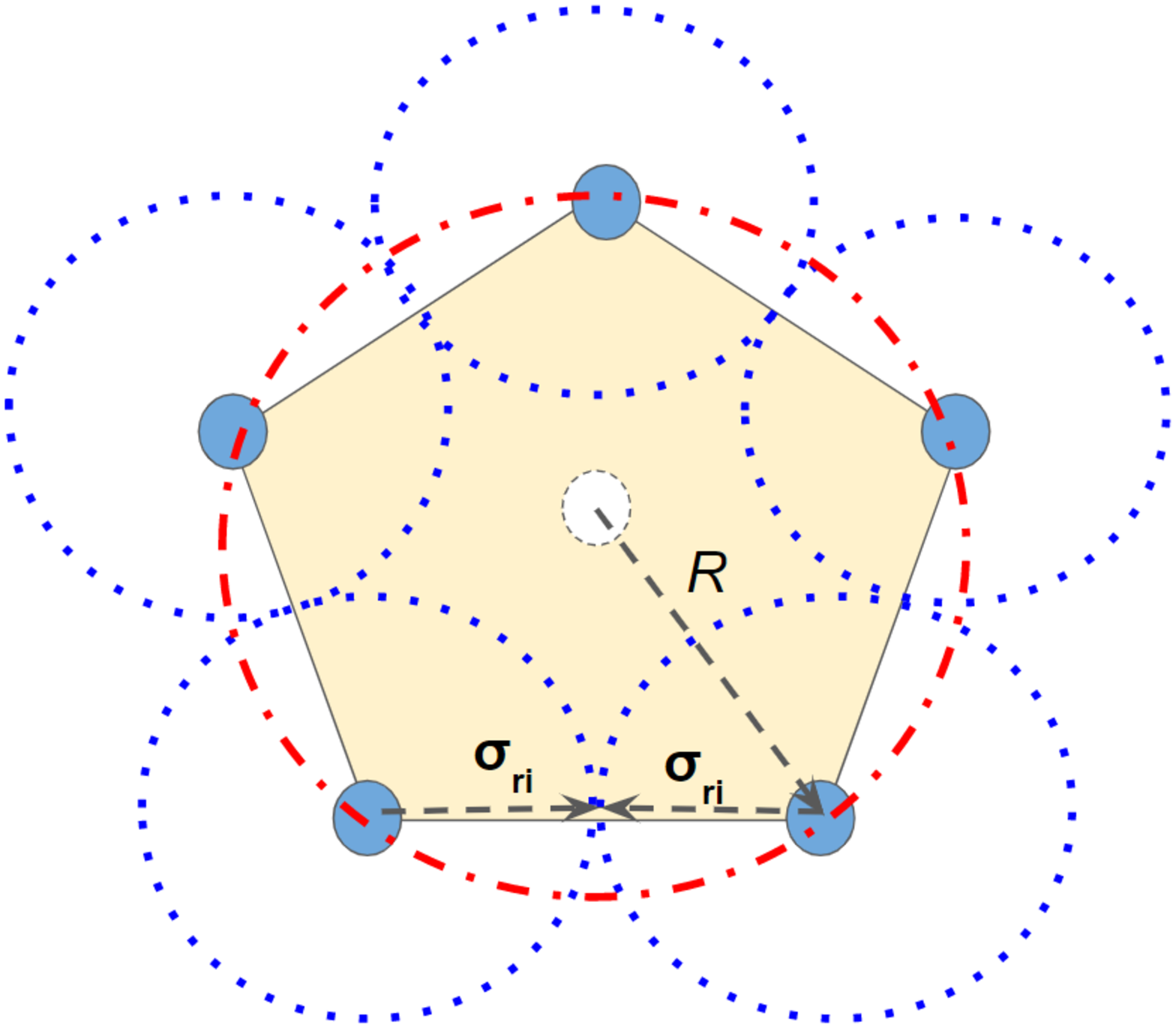}%
\label{F01}}
\caption{Proposed approach - sample schematic.}
\label{fig_sim}
\end{figure}
% \begin{figure}
% \begin{center}
% \includegraphics[width=0.45\textwidth]{sam_vir.png}
%  \caption{Sample virtual function.}
%  \label{F0}
% \end{center}
%  \end{figure} 

 Fig.~\ref{F0} presents a schematic of a FPF with 2 robots $(r_1,r_2)$, starting at two arbitrary locations, moving towards the minimum energy points at a distance \textit{R} from the virtual center. 

For simplicity, this paper focuses on formations that are regular convex polygons. This is achieved by choosing the same repulsive field for each of the $N$ robots. While this simplifies the problem, this limits the method in achieving a desired orientation of the formation. Nonetheless, it will ensure that the robots reach an equilibrium at a distant $R$ from the center of the formation. These equilibrium values would be decided by the number of robots in the formation. A typical polygonal formation using five robots is presented in Fig.~\ref{F01} characterized by the parameters \textit{R}, a user-defined parameter, and $\sigma_{ri}$, decided by the number of robots, $N$. Larger \textit{N} brings the robots closer to each other and may lead to formation instabilities in the presence of perturbations. This could be addressed by redesigning $U_{ij}$ as a function of \textit{N}, but such an approach is beyond the scope of this paper and would be attempted in the future.
% \begin{figure}
% \begin{center}
% \includegraphics[width=0.45\textwidth]{sam_form.png}
%  \caption{A typical formation based on the proposed approach.}
%  \label{F01}
% \end{center}
%  \end{figure} 
\subsection{Candidate FPF function}
To demonstrate the efficacy of the proposed concept, the following candidate function satisfying aforementioned properties is proposed.
\begin{equation}
  U_{v}(\underline{q}_i) = 1+\tanh^2{\sigma_1d_v}-K_v\tanh^2{\sigma_2d_v},
  \label{vir}
\end{equation}
where $\sigma_1$, $\sigma_2$, and $K_v$ are the design parameters, and $d_v = ||\underline{q}_i-\underline{q}_{v}||$. For simplicity, consider the virtual agent is located at the origin, i.e., $\underline{q}_{v} = [0~ 0]^T$.

\subsubsection{Boundary conditions}
\begin{itemize}
    \item 
    \begin{equation}
    \begin{aligned}
      \lim_{\underline{q_i}\to 0} U_{v}(\underline{q_i})  &= 1+ \lim_{\underline{q_i}\to 0}\tanh^2{\sigma_1||\underline{q}_i||} \\&- K_v\lim_{\underline{q_i}\to 0}\tanh^2{\sigma_2||\underline{q}_i||} = 1.
      \end{aligned}
    \end{equation}
    \item 
    \begin{equation}
    \begin{aligned}
      \lim_{\underline{q_i}\to \infty} U_{v}(\underline{q_i})  &= 1+ \lim_{\underline{q_i}\to \infty}\tanh^2{\sigma_1||\underline{q}_i||} \\&- K_v\lim_{\underline{q_i}\to \infty}\tanh^2{\sigma_2||\underline{q}_i||}\\
      &=1 + 1 - K_v = 2-K_v.
      \end{aligned}
    \end{equation}
\end{itemize}
Since global maximum is assumed to be at $\underline{q}_{v} = [0~0]^T$, the value of the function $U_{v}(\underline{q}_i)$ at global maximum is $\lim_{\underline{q}_i\to 0} U_{v}(\underline{q}_i)=1$.

Hence, 
\begin{equation}
\begin{aligned}
        &\lim_{\underline{q}_i\to \infty} U_{v}(\underline{q}_i) = 2-K_v < 1\\
        &\Rightarrow K_v>1.
        \label{Kv}
\end{aligned}
\end{equation}
\subsubsection{Relation between parameters $\sigma_1$, $\sigma_2$, and $K_v$}

Let the force acting on individual robot be,
\begin{equation}
\begin{aligned}
    F_v(\underline{q}_i) &= -\nabla_i U_{v}(\underline{q}_i),\\
    &=-\frac{\partial U_{v}(\underline{q}_i)}{\partial \underline{q}_i},\\
   &= -\frac{\partial}{\partial \underline{q}_i} \Big(1+\tanh^2{\sigma_1||\underline{q}_i||}-K_v\tanh^2{\sigma_2||\underline{q}_i||}\Big).\\
   &=\Bigg(-\frac{2\sigma_1\tanh{\sigma_1||\underline{q}_i||}}{\cosh^2{\sigma_1||\underline{q}_i||}}+\frac{2K_v\sigma_2\tanh{\sigma_2||\underline{q}_i||}}{\cosh^2{\sigma_2||\underline{q}_i||}}\Bigg)\frac{\underline{q}_i}{||\underline{q}_i||}.
\end{aligned}  
\label{Fv}
\end{equation}
At equilibrium point, $F_v(\underline{q}_{ei})=0$ and
there exists multiple equilibrium points corresponding to, $\underline{q}_{e1i}=0$, $\frac{\sigma_1\tanh{\sigma_1||\underline{q}_{e2i}||}}{\cosh^2{\sigma_1||\underline{q}_{e2i}||}}=\frac{K_v\sigma_2\tanh{\sigma_2||\underline{q}_{e2i}||}}{\cosh^2{\sigma_2||\underline{q}_{e2i}||}}$, and  $\infty$.

The virtual centre is assumed to be located at $\underline{q}_{e1i}=0$, and this corresponds to the global maximum of the FPF. Now, applying second derivative test, for a maximum,
\begin{equation}
    F'_v(\underline{q}_i)|_{\underline{q}_{e1i}=0}=\frac{\partial F_{v}(\underline{q}_i)}{\partial \underline{q}_i}|_{\underline{q}_{e1i}=0}<0.
\end{equation}

\begin{equation}
\begin{aligned}
    F'_v(\underline{q}_i)|_{\underline{q}_{e1i}=0}&=
    \Bigg(\frac{2K_v\sigma_2^2-4\sigma_2^2\sinh^2{\sigma_2||\underline{q}_{e1i}||}}{\cosh^4{\sigma_2||\underline{q}_{e1i}||}}\\&-\frac{2\sigma_1^2-4\sigma_1^2\sinh^2{\sigma_1||\underline{q}_{e1i}||}}{\cosh^4{\sigma_1||\underline{q}_{e1i}||}}\Bigg)\frac{\underline{q}^2_{e1i}}{||\underline{q}_{e1i}||^2}<0,\\
   &= \sigma_1^2-K_v\sigma_2^2<0,\\&=
   \frac{\sigma_1}{\sigma_2}<\sqrt{K_v}, ~ K_v>1.
   \label{sigma}
\end{aligned}
\end{equation}
This expression gives a criterion for the design of the FPF.

It is not possible to solve $\frac{\sigma_1\tanh{\sigma_1||\underline{q}_{e2i}||}}{\cosh^2{\sigma_1||\underline{q}_{e2i}||}}=\frac{K_v\sigma_2\tanh{\sigma_2||\underline{q}_{e2i}||}}{\cosh^2{\sigma_2||\underline{q}_{e2i}||}}$ analytically and to find the values of $\underline{q}_{e2i}$. The equilibrium point $\underline{q}_{e2i}$ represents the intersection points of the left and right sides of the aforementioned expression, and is essentially the radius of the formation, \textit{R}.

\begin{figure}
\begin{center}
\includegraphics[width=0.45\textwidth]{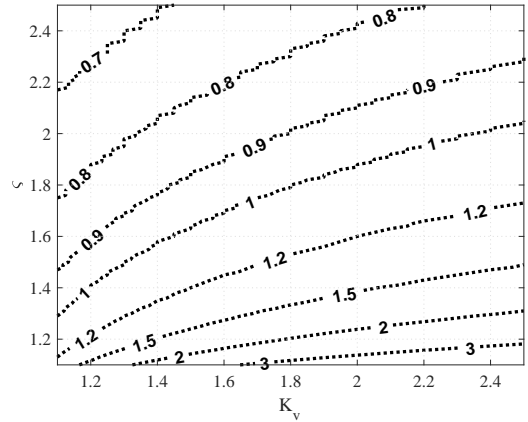}
 \caption{Contour plots presenting $\mathcal{R}$ values for different $K_v$ and $\varsigma$ values.}
 \label{F02}
\end{center}
 \end{figure}

If,  
\begin{equation}
    \frac{\sigma_1\tanh{\sigma_1||\underline{q}_{e2i}||}}{\cosh^2{\sigma_1||\underline{q}_{e2i}||}}-\frac{K_v\sigma_2\tanh{\sigma_2||\underline{q}_{e2i}||}}{\cosh^2{\sigma_2||\underline{q}_{e2i}||}}=0,
\end{equation}
and on substituting $\sigma_1||\underline{q}_{e2i}||=\mathcal{R}$, and $\sigma_2/\sigma_1=\varsigma$ gives,
\begin{equation}
    \frac{\tanh{\mathcal{R}}}{\cosh^2{\mathcal{R}}}-\frac{K_v\varsigma \tanh{\varsigma \mathcal{R}}}{\cosh^2{\varsigma \mathcal{R}}}=0.
    \label{num}
\end{equation}
By numerically solving Eq.~\ref{num} for different $K_v$ and $\varsigma$ values imposing constraints from Eq.~\ref{Kv} and Eq.~\ref{sigma}, it is possible to get a design map for the selection of parameter values for different values of scaled formation radius, $\mathcal{R}$. From the above relations, one could notice that the solutions exist only for $(K_v,\varsigma)>1$. A sample map for $1<K_v\leq 2.5$, $1<\varsigma \leq 2.5$ is presented in Fig.~\ref{F02}. From this map, for a formation of interest, assuming a fixed $\sigma_1$, it is possible to obtain the parameters for the proper design of the FPF as presented in Eq.~\ref{vir}. As suggested by Eq.~\ref{sigma}, $K_v$ and $\varsigma$ follow an inverse relationship. For lower values of $K_v$, the formation would span more area for higher values of $\varsigma$, and vice versa. Also, as $K_v$ increases, the maximum attainable formation size reduces considerably. One can use similar set of maps for designing a formation of their choice of \textit{R}.

 \subsubsection{Monotonicity }
In order to prove the function $U_{v}(\underline{q}_i)$ is monotonically decreasing in the open interval ($\underline{q}_{v},\underline{q}_{R}$), $\underline{q}_{R}$ being $R$ units away from $\underline{q}_{v}$, it sufficient to prove $F_v(\underline{q}_i) = -\nabla_i U_{v}(\underline{q}_i)<0$ in that interval. 

For the considered FPF, from Eq.~\ref{Fv}, $F_v(\underline{q}_i)<0$,
\begin{equation}
   \Rightarrow \sigma_1\tanh{\sigma_1||\underline{q}_i||}-K_v\sigma_2\tanh{\sigma_2||\underline{q}_i||}<0.
\end{equation}
From Eq.~\ref{Kv} and Eq.~\ref{sigma}, $K_v>1$, and $\sigma_2>\sigma_1$, and by using properties of hyperbolic functions, $\sigma_1\tanh{\sigma_1||\underline{q}_i||}<K_v\sigma_2\tanh{\sigma_2||\underline{q}_i||}$ and $F_v(\underline{q}_i)<0$. Hence the function $U_{v}(\underline{q}_i)$ is monotonically decreasing in the interval ($\underline{q}_{v},\underline{q}_{R}$). Similarly, it is possible to prove the monotonicity in the interval ($\underline{q}_{R},\infty$), and is omitted for brevity. This condition eliminates the existence of local minima all along the field, which would trap the agents and prevents forming a convex polygonal structure around the virtual agent.

 \begin{figure*}[h]
\centering
\subfloat[Case I - Five agents.]{\includegraphics[width=0.5\textwidth]{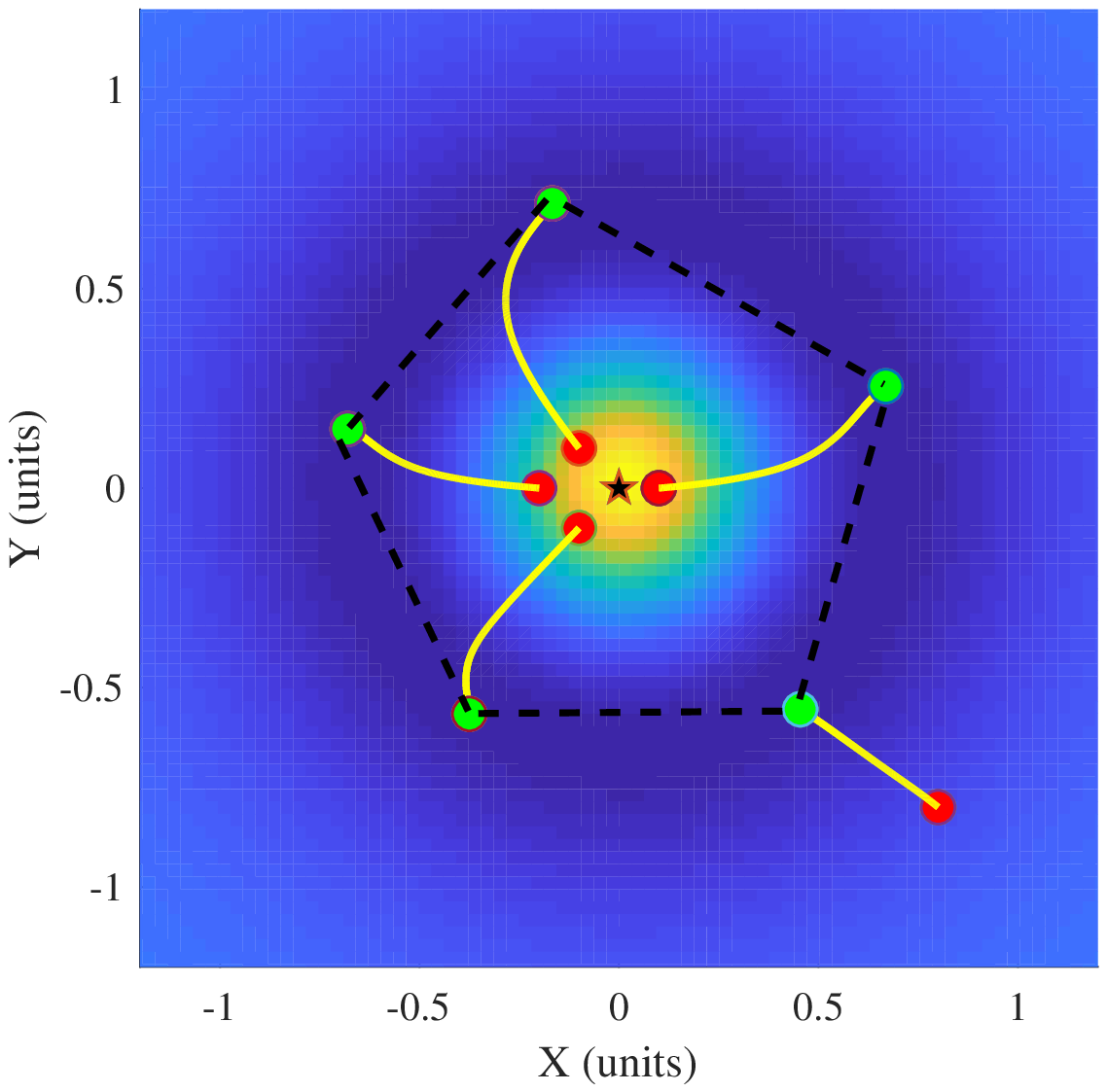}%
\label{fig_first_case}}
%\hfil
\subfloat[Case II - Ten agents.]{\includegraphics[width=0.5\textwidth]{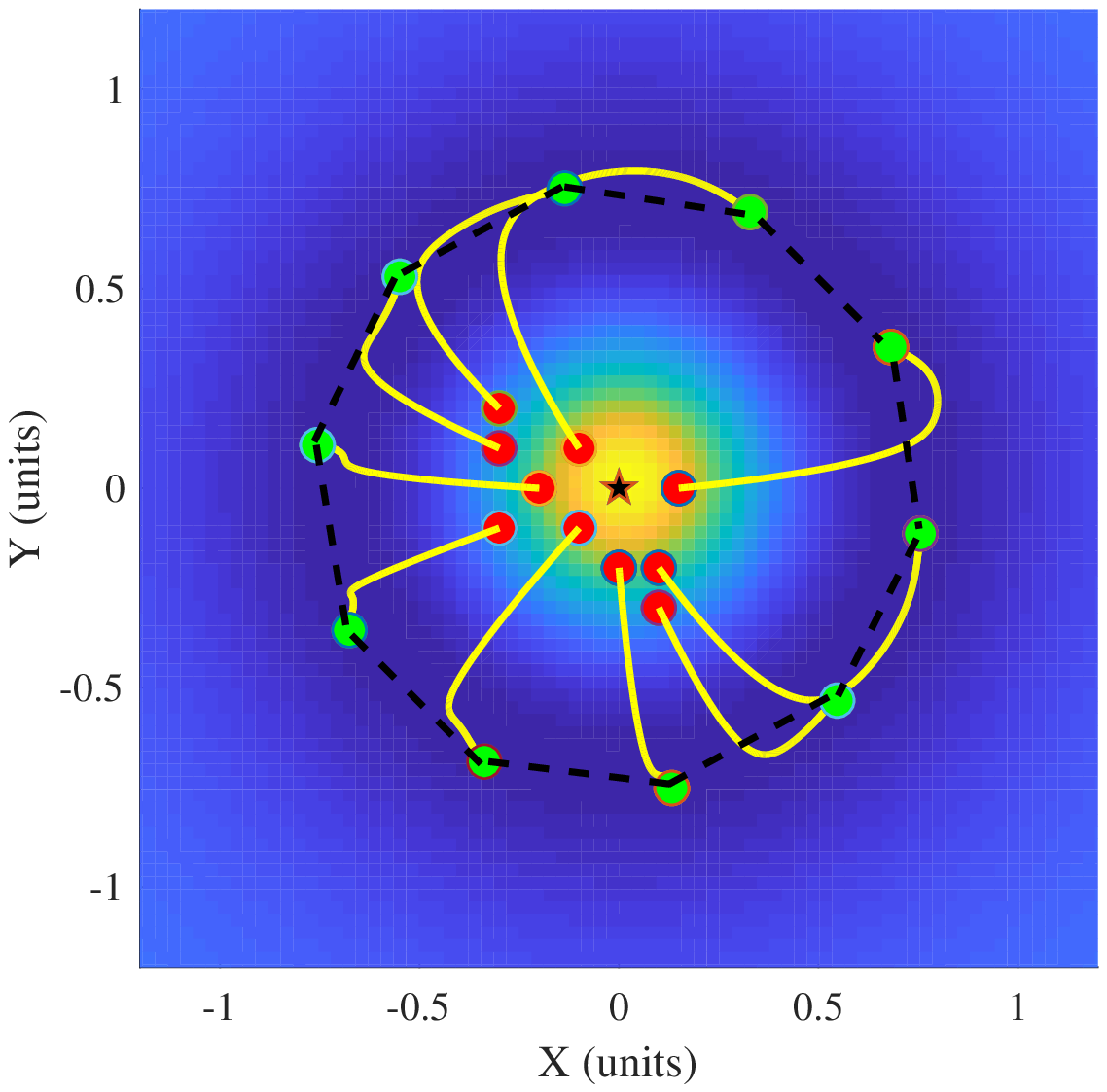}%
\label{fig_second_case}}
\caption{Surface plots showing agents in polygonal formation with $\mathcal{R}=0.8$ }
\label{fig_sim}
\end{figure*}

 \section{Path planning}
 \subsection{Virtual Agent Path Planning}
 Consider \textit{N} robots starting from distinct arbitrary locations in a configuration space, $\mathcal{CS}$, establishing a formation around a virtual agent as presented in Section II. Then, starting from an initial configuration corresponding to the virtual agent location $\underline{q}_{v} = \underline{q}_{I}$, a collision-free path is planned in the free configuration space, $\mathcal{FS}$ to the goal configuration $\underline{q}_{v} = \underline{q}_{G}$. At any instant, the virtual agent is assumed to be under the influence of the attractive potential from the goal and repulsive potential from the obstacles.
 \begin{equation}
     U_{va} = U_{G} + U_{obs},
 \end{equation}
 where $U_{va}$ is the total potential experienced by the virtual agent, $U_{G}$ is the attractive potential from the goal, and $U_{obs}=\sum_{k = 1}^{M} U_{k}$ represents the total repulsive potential from \textit{M} obstacles in the field. 
 Attractive potential field is selected as,
 \begin{equation}
     U_{G} = \frac{1}{2}\lambda d_{vg}^2,
 \end{equation}
 where $d_{vg} = ||\underline{q}_{v}-\underline{q}_{G}||$, and $\lambda$ is a positive constant. The repulsive field due to an individual obstacle is modelled as an exponential function,
 \begin{equation}
     U_k = K_r \exp(-\sigma_od_{vk}),
 \end{equation}
 where $d_{vk} = ||\underline{q}_{v}-\underline{q}_{k}||$, $\underline{q}_k$ represents the location of $k^{th}$ obstacle, $\sigma_o$ is a positive exponent representing the potential field spread, and $K_r$ is a positive constant.
 
 Total force on the virtual agent is given by,
 \begin{equation}
 \begin{aligned}
    F_{va}(\underline{q}_{v}) &= -\nabla U_{va}
    =-\lambda (\underline{q}_{v}-\underline{q}_{G}) \frac{(\underline{q}_{v}-\underline{q}_{G})}{||\underline{q}_{v}-\underline{q}_{G}||}\\& + \sum_{k = 1}^{M} K_r\sigma_o\exp(-\sigma_od_{vk})\frac{(\underline{q}_{v}-\underline{q}_{k})}{||\underline{q}_{v}-\underline{q}_{k}||}. 
    \label{fva}
 \end{aligned}
 \end{equation}
Now, if one were to consider $F_{va}(\underline{q}_{v})$ as generalized accelerations, $\ddot{\underline{q}}_{v} = F_{va}(\underline{q}_{v})$, then it is possible to numerically integrate and find the collision-free virtual agent path $\underline{q}_{v}$. It should be noted that velocity damping is necessary to achieve absolute stability, and hence is included for simulation studies.
% \begin{equation}
% \begin{aligned}
%   &\underline{v}_{{vir}_{k+1}} = \underline{v}_{{vir}_k} + \delta F_{va}(\underline{q}_{{vir}_k}),\\
%   & \underline{q}_{{vir}_{k+1}} = \underline{q}_{{vir}_k} + \delta \underline{v}_{{vir}_k}.
% \end{aligned}
% \end{equation}

\subsection{Path Planning for Individual Robots}
The individual robot paths are planned such that they closely follow a desired formation around the virtual agent as presented in Section III. The total potential on each individual robot is a combination of the virtual agent potential, inter-robot repulsive potentials and obstacle repulsive potentials, and is given by,
\begin{equation}
    U_i = U_{v} + \sum_{i \neq j}^{N} U_{ij} + \sum_{k = 1}^{M} U_{ik}.
    \label{tp}
\end{equation}
 The goal potential is not affecting $U_i$ explicitly, but is indirectly driving the robots towards the goal by attracting the virtual agent towards the goal. For each $\underline{q}_{v}$ generated using Eq. \ref{fva}, a new set of minima are generated, and the robots are forced to move to these new locations. As presented in Eq.~\ref{tp}, in the neighborhood of the $k^{th}$ obstacle, its repulsive potential $U_{ik}$ affects the total potential experienced by individual robots, changing the minimum value, and thus their positions. For a robot $i$ that is closer to the obstacle, any change in its position bring it closer to the other robots, thus creating a new minima for other $N-1$ robots.   This results in a new configuration of the formation. Once the influence of the obstacle potential decays, the robots return back to their original polygonal formation dictated by the FPF.

\section{Results and Discussions}

%  \begin{figure}[h]
% \begin{center}
% \includegraphics[width=0.5\textwidth]{vc.eps}
%  \caption{Contour plots showing six agents in formation with $\mathcal{R}=0.8$.}
%  \label{F03}
% \end{center}
%  \end{figure} 
Simulations are conducted for the sample FPF as presented in Eq. (3) with $K_v = 2$, $\sigma_1 = 1$, and $\sigma_2 = 2.4$ (such that, $\varsigma = 2.4$). The FPF corresponding to these parameter values is assumed to be having a peak at $\underline{q}_{v}=(0,0)$. The peak is surrounded by a circle of minima with $\mathcal{R}=0.8$. This value of $\mathcal{R}$ could be verified from the contour plot presented in Fig.~\ref{F02}. The function is designed according to the conditions presented in Section II to rule out existence of any local minima and to ensure the robots converge around the virtual agent to form a convex polygon.

  \begin{figure*}[h]
\begin{center}
\includegraphics[width=0.8\textwidth]{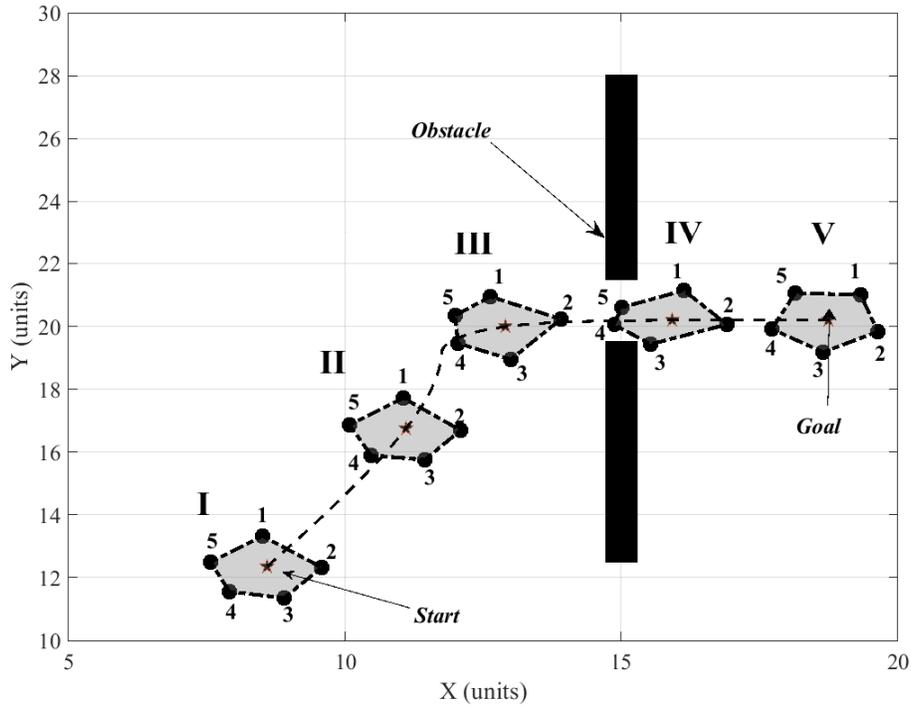}
 \caption{Plots showing the flexible formation in action.}
 \label{F05}
\end{center}
 \end{figure*} 
 
Figure \ref{fig_sim} shows the surface plots of agents starting at arbitrary locations coming together to polygonal formations around the virtual agent. Two distinct cases with five and ten agents forming a symmetric pentagon and decagon utilizing the proposed design methodology are presented in Figs.~\ref{fig_first_case} and \ref{fig_second_case}, respectively. The virtual agent/function location is marked by a pentagram, red and green round markers represent the starting and final location of the agents, respectively. The formation is established at $\mathcal{R}=0.8$ around the virtual agent. The FPF moves the robots radially towards the minima, and the individual robot repulsive fields push the robots away from each other in tangential direction. Once they reach an equilibrium, they would remain at minima as the vertices of a convex polygon, as shown in Fig.~\ref{fig_sim}.

After getting into the formation, the next goal is to plan a collision-free path to reach the goal following the procedure presented in Section IV. A collision-free path is computed for the virtual agent from the initial configuration to the goal configuration. A quadratic potential attracts the virtual agent towards the goal, and an exponential potential prevents it from colliding with the obstacles. The virtual agent would be following a minimum potential path to the goal, and plots showing the same are omitted for brevity. 

The agents are not affected by the goal potential directly, but is forced to follow the virtual agent towards the goal configuration. Agents, under the effect of external forces imposed by the obstacle potentials should be able to move in the obstacle ridden environment without colliding with the obstacle and each other, while maintaining a formation around the virtual agent. As the formation move away from the influence of the obstacle field, the deformed polygonal structure should be reverted back to its former-self.

Figure \ref{F05} presents the proposed approach in action. A pentagonal formation of five robots (as presented in Fig.~\ref{fig_first_case}) is assumed to be moving towards a goal through a narrow pathway. Five distinct configurations, I-V are used to present the state of the formation at different instances. Configuration I is assumed to be the starting configuration, and in both configurations I and II, the robots are far away from the obstacle field and are not under its influence. So, at each step, the agents are trying to maintain a regular pentagonal formation around the virtual agent. Once they reach closer to the obstacle, under the influence of its repulsive field, the robots would try to move farther away thus avoiding collisions. But, these robots are still under the influence of the virtual agent potential and they are bound to stay together, without colliding with each other. Under these conditions, the robots would now try to find new equilibria around the virtual agent, but now with higher energy values, accommodating the influence of the obstacle repulsive field. This would cause deformations in the formation shape as presented in configurations III and IV. Even after passing through the narrow passage as shown in configuration IV, an individual robot is not completely free from the influence of the obstacle field. The robots, that are still under the influence of the obstacle fields, i.e., those that are still passing through the passage have a say on the forces that are influencing the robot motion. The formation would be forced to \textit{squeeze} through the pathway as shown in configuration III. Once the whole formation is away from the influence of the obstacle field, they revert back to their minimum energy equilibrium positions corresponding to the symmetric pentagonal structure as illustrated in configuration V.
 \begin{figure}[!h]
\begin{center}
\includegraphics[width=0.45\textwidth]{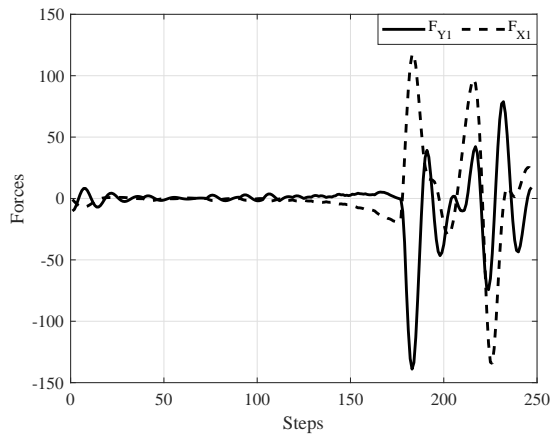}
 \caption{Forces acting on robot 1 in Fig.~\ref{F05}.}
 \label{F06}
\end{center}
 \end{figure} 

The forces acting on an individual robot (robot 1) is presented in Fig.~\ref{F06}. During the formation assembling phase (as presented in Fig.~\ref{fig_first_case}), the force components in both X and Y directions are nonzero, forcing the robot to move towards the minima. Once in formation, then as the virtual agent moves, the robot would be dragged along with less effort, indicated by close-to-nil forces magnitudes in Fig.~\ref{F06} (configurations I and II). Once the formation reaches near the obstacle, robot 1, being closest to the obstacle, would have to deviate more from its vertex than other agents. This would need forces of higher magnitude, since the forces have to overcome the virtual agent forces and other agent repulsive forces to move from its equilibrium vertex point. As shown in Fig.~\ref{F06}, first a net negative force is applied in the Y-direction to make it move away from the obstacle and towards the virtual agent, after that, the robot is pushed towards right to push out of the obstacle vicinity. This force distribution would be different on different robots, and this makes the formation flexible and makes it quasi-centralized. 
\section{Conclusion}
This paper presents a novel \textit{quasi-centralized} approach to address the path planning problem of multi robot systems in an obstacle ridden environments. In this approach, a centralized planner generates the path for a \textit{virtual agent} located at the center of the formation. Subsequently, the paper introduces the concept of formation potential fields (FPF) for self-organization into a formation around the virtual agent. Additionally, the FPF will ensure the robots to stay in formation during navigation. In the neighborhood of obstacles, the robots selfishly plan a collision-free path to avoid the obstacles, all the while closely maintaining the formation. The approach of combining centralized and decentralized path planning approaches made the formation flexible and allowed the multi-robot system to have a collision-free navigation. 

The paper details the characteristics of class of FPFs and proposes a parameterized candidate function. Further, a detailed guide is given for  selection of these FPF parameters to realize formations of different radii.  The simulation results show that the quasi-centralized approach is effective in planning obstacle avoidance paths for MRS with minimal deviation from the formation. In comparison to other approaches that scale, split or fully deviate the formation, this approach is expected to minimize the work done by the formation. However, more formal analysis to this extent is needed and is beyond the scope of the current paper. Further, the proposed algorithm does not allow for desired orientations of the formation and this becomes part of our future work.

\section*{Acknowledgements}

The authors acknowledge discussions with Prof. Chetan Pahlajani, Faculty in Discipline of Mathematics, IIT Gandhinagar and Mr. Aditya M Rathi, Project Assistant in SMART Lab for their valuable discussion. The authors also acknowledge the support of Indian Institute of Technology Gandhinagar.


% \section*{Acknowledgment}




\bibliographystyle{IEEEtran}
\bibliography{bare_conf.bib}

% \bibitem{1}

% \end{thebibliography}

\end{document}